\documentclass[letterpaper, 10 pt, conference]{ieeeconf}  
\usepackage{multirow}
\usepackage{amsmath}
\usepackage{array}
\usepackage{tabu}
\usepackage{graphicx}
\usepackage{amsmath}
\usepackage{scalerel}
\usepackage{amssymb}
\usepackage{float}

\IEEEoverridecommandlockouts                              
\title{\LARGE \bf
Continuous Gesture Recognition from sEMG Sensor Data with Recurrent Neural Networks and Adversarial Domain Adaptation 
}

\author{Ivan Sosin, Daniel Kudenko, and Aleksei Shpilman
\thanks{I. Sosin is with JetBrains Research,
        St Petersburg, Russia
        {\tt\small iasawseen@gmail.com}}
\thanks{D. Kudenko is with the Department of Computer Science, University of York, UK, the National Research University Higher School of Economics, St Petersburg, Russia, and JetBrains Research,
St Petersburg, Russia
{\tt\small daniel.kudenko@york.ac.uk}}
\thanks{A. Shpilman is with JetBrains Research, St Petersburg, Russia 
and National Research University Higher School of Economics, St Petersburg, Russia
        {\tt\small aleksei@shpilman.com}}
}

\begin{document}

\maketitle
\thispagestyle{empty}
\pagestyle{empty}

\begin{abstract}

Movement control of artificial limbs has made big advances in recent years. New sensor and control technology enhanced the functionality and usefulness of artificial limbs to the point that complex movements, such as grasping, can be performed to a limited extent. To date, the most successful results were achieved by applying recurrent neural networks (RNNs). However, in the domain of artificial hands, experiments so far were limited to non-mobile wrists, which significantly reduces the functionality of such prostheses. In this paper, for the first time, we present empirical results on gesture recognition with both mobile and non-mobile wrists. Furthermore, we demonstrate that recurrent neural networks with simple recurrent units (SRU) outperform regular RNNs in both cases in terms of gesture recognition accuracy, on data acquired by an arm band sensing electromagnetic signals from arm muscles (via surface electromyography or sEMG). Finally, we show that adding domain adaptation techniques to continuous gesture recognition with RNN improves the transfer ability between subjects, where a limb controller trained on data from one person is used for another person.

\end{abstract}

\section{Introduction}

One of the prominent fields in robotics is the development of artificial or prosthetic limbs. The most advanced devices are controlled via electromagnetic sensors that measure sEMG signals either from muscles \cite{fajardo2015} or the central nervous system \cite{Li2017}. 

Artificial limbs are normally controlled via gesture recognition from the sEMG sensor data. This gesture recognition can be either discrete \cite{Du2017} or continuous \cite{Quivira2018}. In the discrete case, there is a limited number of available gestures that the limb can perform, and the goal is to map the sensor data to one of these gestures. Therefore, the recognition task is a classification problem. In our work, we focus instead on continuous gesture recognition, which derives the angles of the limb joints from the data and allows for more life-like limb movement. 

In this paper, we perform experiments on the task of controlling a robotic hand via an arm band sensor. Previous work (e.g. \cite{fajardo2015, Du2017, Quivira2018}) assumed robotic hands with immobile wrists, which  significantly restricts the functionality of the robotic limb. We present, for the first time, experiments with both immobile and mobile wrist data, and show that our proposed techniques outperform the state-of-the-art in both cases. 

The most successful gesture recognition from the sEMG sensor data methods to date employ recurrent neural networks (RNNs) \cite{Quivira2018}. However, RNNs take a very long time to train to reach a desirable accuracy \cite{Lei2017}. In our work, we propose to use RNNs with simple recurrent units (SRU)  \cite{Lei2017}, instead, for continuous gesture recognition. This method has the advantage of reaching a higher gesture recognition accuracy than regular RNNs in the same training time.

In addition, we show how domain adaptation techniques \cite{Ganin2014} can be used for continuous gesture recognition, and for the first time empirically demonstrate that these techniques improve the transferability between subjects, where a limb controller trained on data from one person is used for gesture recognition of another person.

\section{Background and Related Work}

\subsection{sEMG sensors}
\begin{figure}[thtb]
    \centering
    \includegraphics[width=0.45\textwidth]{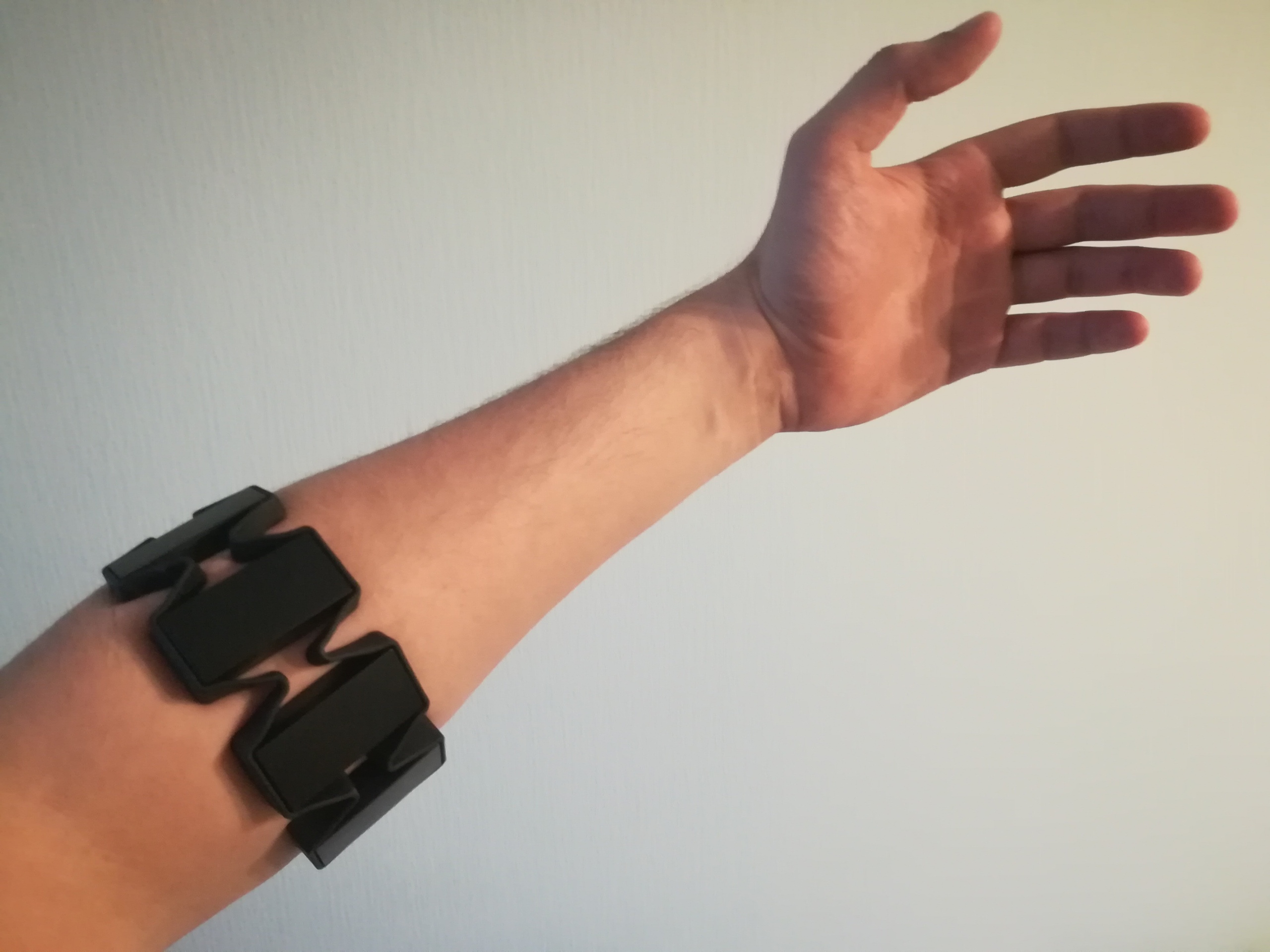}
    \caption{Myo Armband}
    \label{img:myo}
\end{figure}

Surface electromyography is the method of obtaining electric signals coming from neurons that are responsible from muscle contractions. As the name suggests, it is a non-invasive method and electrodes are attached to the skin surface. These signals can be then used in a control system for powered upper-limb prostheses because motions of wrist and fingers are mostly controlled by muscles in the forearm \cite{phinyomark2012}.
sEMG sensors range from non-portable units with high measurement accuracy to light and wearable units, albeit with a lower accuracy.
In order to retrieve sEMG data, we utilized the Myo armband \cite{Sathiyanarayanan2016}, as shown in Figure~\ref{img:myo}. This device has the advantages of being wearable and having an affordable price\footnote{At the time of writing the price is in the range of \$200}. The Myo armband consists of 8 sEMG sensors capable of recording sEMG data at frequency of 200 Hz. The raw signal is digitized with 8-bit analog-to-digital converter to yield a $[-128,+128]$ range. For wireless connection the Myo armband uses Bluetooth Low Energy technology.

\subsection{Motion capture system}

In order to tackle the task of continuous gesture recognition, a motion capture system to extract the limb angles is required for the generation of the data set. Existing motion capture systems are based on a wide range of technologies, including video recording \cite{kim2008}, ultrasound \cite{hettiarachchi2015}, and tracking gloves \cite{kim2009}.

In our work, the Leap Motion Controller \cite{leapmotion} was used, which provided a cheap and efficient way to track and record the limb angles of a user's hand. This device is specifically designed for hand movement detection. The controller itself consists of two monochromatic IR cameras and three IR LEDs. The overall accuracy of the controller was shown to be 0.7 millimeters. Motion capture data is transmitted with a frequency from 50 Hz up to 120 Hz.



\subsection{Recurrent Neural Networks}
Recurrent neural networks (RNNs) \cite{Jain:1999:RNN} are a special type of neural networks designed to work with (numerical) sequence data. The input of an RNN is the sequence and the output is a numerical vector. In our case, the input sequence are the sEMG measurements, and the output is a vector of limb angles. Every neuron of the network on each element of the input sequence updates its state and outputs a value as defined by:
\begin{equation}
    h_t = \sigma_h(W_h x_t + U_h h_{t - 1} + b_h) \\
\end{equation}
\begin{equation}
    y_t = \sigma_y(W_y h_t + b_y)
\end{equation}
where $x_t$ is the sequence element, $h_t$ is the hidden state, $y_t$ is the output vector, $W, U, b$ are cell parameters, and $\sigma_h$ and $\sigma_y$ are activation functions.

\subsection{Hand Gesture Recognition}

Initial work on controlling artificial limbs using EMG signals focused on creating a control scheme, i.e. mapping specific signals to specific actions of the limb (e.g. \cite{fajardo2015}). These approaches are not able to tackle a large variety of gestures and the resulting controllers are specific to an individual and can not be transferred to other persons. Gesture recognition methods have been introduced to overcome these limitations. 

The gesture recognition task can be split into:
\begin{enumerate}
    \item Classification with a number of discrete gestures or poses.
    \item Regression with respect to recognizing continuous values such as joint angles.
\end{enumerate}

In most recent and advanced work, Du et al.\ \cite{Du2017} propose a hand gesture classification method employing domain adaption (specifically adaptive batch normalization, AdaBN). The authors generated several datasets with the number of gestures varying from 8 to 12, none of which involved wrist movement. The results showed that AdaBN can improve generalization and tranferability of a convolutional neural network with respect to accuracy on the discrete gesture recognition tasks.
Following up on this work, \cite{Quivira2018} focused on continuous gesture recognition (joint angles) using Gaussian processes. Again, unlike our work, this research did not involve any wrist movement.

\section{Method}

In this section, we present our approach for continuous gesture recognition. Specifically, we propose the use of RNNs with simple recurrent units (SRUs) to achieve a higher accuracy in the same time of training than regular RNNs using gated recurrent units (GRUs). Furthermore, we introduce a novel technique to apply domain adaptation to continuous gesture recognition. 

\subsection{Simple Recurrent Units}

Due to vanishing gradient and exploding gradient problems vanilla RNN is not used nowadays \cite{chung2014}. The most common approach to date is to employ RNNs with GRU cells.

\subsubsection{GRU-cell}
The GRU-cell features special gates to reduce vanishing and exploding gradient problems. It is defined as: 
\begin{equation} \label{eq:gru_z_t}
    z_t = \sigma_g (W_z x_t + U_z h_{t - 1} + b_z)
\end{equation}
\begin{equation} \label{eq:gru_r_t}
    r_t = \sigma_g (W_r x_t + U_r h_{t - 1} + b_r)
\end{equation}
\begin{equation}
    h_t = (1 - z_t) \odot h_{t - 1} + z_t \odot \sigma_h (W_h x_t + U_h (r_t \odot h_{t - 1}) + b_h)
\end{equation}

where $x_t$ is the sequence element, $h_t$ is the hidden state and output, $z_t$ is the update gate, $r_t$ is the reset gate, and $W, U, b$ are cell parameters.

\begin{figure}[thtb]
    \centering
    \includegraphics[width=0.40\textwidth]{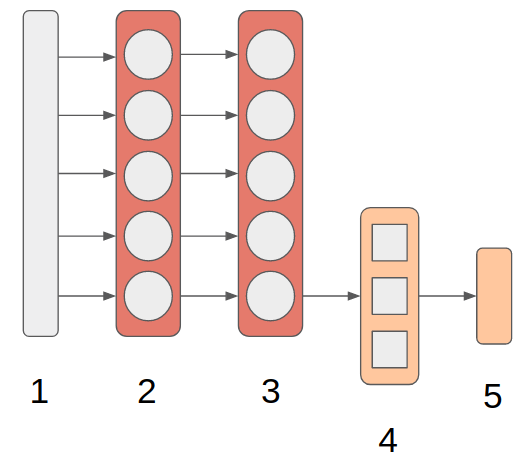}
    \caption{Recurrent neural network architecture \newline 1 - Input data, \newline 2 - The first GRU/SRU layer, \newline 3 - The second GRU/SRU layer, \newline 4 - Fully-connected layer of Predictor, \newline 5 - Fully-connected layer without activation-function - result vector.}
    \label{img:rec_scheme_num}
\end{figure}

Figure~\ref{img:rec_scheme_num} illustrates an RNN architecture with GRU cells.

\subsubsection{SRU-cell}
The GRU-cell, due to its internal dependencies, is difficult to parallelize efficiently. In order to reduce computation cost for RNN training and inference, a simple recurrent unit was proposed in \cite{Lei2017} as follows:

\begin{equation}
    \hat{x_t} = W x_t
\end{equation}
\begin{equation}
    f_t = \sigma(W_f x_t + b_f)
\end{equation}
\begin{equation}
    r_t = \sigma(W_r x_t + b_r)
\end{equation}
\begin{equation}
    c_t = f_t \odot c_{t - 1} + (1 - f_t) \odot \hat{x_t}
\end{equation}
\begin{equation}
    h_t = r_t \odot \sigma(c_t) + (1 - r_t) \odot x_t
\end{equation}

where $x_t$ is a sequence element, $h_t$ is the output vector, $c_t$ is the hidden state, $f_t$ is the forget gate, $r_t$ is the reset gate, and $W, U, b$ are cell parameters.

The RNN architecture with SRU cells is analogous to the RNN architecture shown in Figure~\ref{img:rec_scheme_num}.

\subsection{Domain adaptation}

When transferring a gesture recognition network trained on one person to another person, specialized transfer learning techniques are required. Adversarial domain adaptation (ADA) has been shown in recent work \cite{ganin2014unsupervised} to perform well for transfer learning tasks, and we have thus selected it for our research. ADA employs an additional neural network that predicts the current domain of input data from a feature vector. There is a gradient reversal layer between the feature vector and the discriminator network. This layer multiplies gradients by some value between -1 and 0. To train a neural network with ADA, we sum the loss function of the predictor and the loss function of the discriminator and perform back-propagation with a reversal of the discriminator gradient before the feature layer. The goal of this process is to inhibit domain-specific features and thus prevent over-fitting.

\begin{figure}[thtb]
    \centering
    \includegraphics[width=0.40\textwidth]{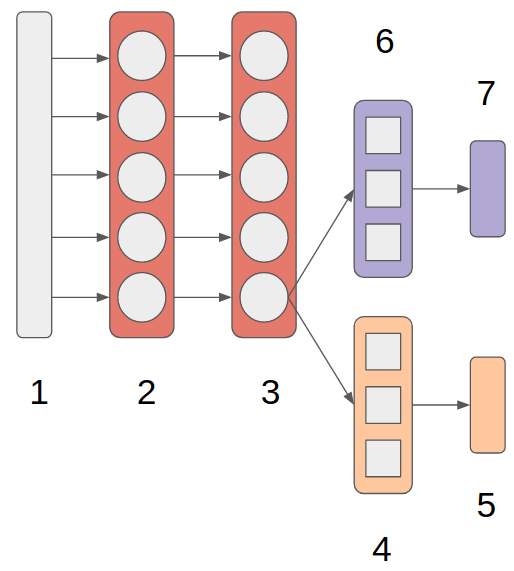}
    \caption{Recurrent neural network with domain adaptation architecture \newline 1 - Input data, \newline 2 - The first GRU/SRU layer, \newline 3 - The second GRU/SRU layer, \newline 4 - Fully-connected layer of Predictor, \newline 5 - Fully-connected layer without activation-function - result vector, \newline 6 - Fully-connected layer of Discriminator, \newline 7 - Fully-connected layer without activation function - vector of classes (domains) probabilities.}
    \label{img:rec_scheme_adapt}
\end{figure}

Figure~\ref{img:rec_scheme_adapt} shows the RNN architecture with domain adaptation. 

We have also tried adaptive batch normalization, as used for discrete hand gesture recognition in \cite{Du2017}, but this did not show good performance on our continuous gesture recognition tasks.

\section{Data acquisition}

sEMG data was obtained by means of the Myo armband device and finger and wrist angles were captured with the Leap Motion motion capture system.

We recorded data recorded from 5 healthy volunteers ranging in age from 20 to 28 years. Each participant was asked to sit on a chair and establish a relaxed position with the wrist slightly above the Leap Motion sensor to ensure accurate motion capture. The participant was then asked to perform a given sequence of hand movements such as bending or stretching the index finger.   

Two datasets were collected: the first dataset only included movement of fingers with a static wrist (similar to \cite{Quivira2018}, and the second one was aimed to add wrist movements and some casual gestures. Data was collected in sessions with a duration of 4 minutes each. The specifics of the datasets are as follows. 

\hfill

\noindent
Immobile wrist dataset:

\hfill

\begin{enumerate}
    \item Movements of distinct fingers, 150 s
    \item Simultaneous movement of all fingers, 60 s
    \item Free finger movements, 30 s
\end{enumerate}

\hfill

\noindent
Mobile wrist dataset:

\hfill

\begin{enumerate}
    \item Movements of distinct fingers, 60 s
    \item Simultaneous movement of all fingers, 30 s
    \item Pinch, 30 s
    \item Open palm movements, 30 s
    \item Thumb movements, 30 s
    \item Free finger movements, 30 s
\end{enumerate}

\hfill

For each volunteer 8 sessions per dataset were recorded. For the mobile wrist dataset, the volunteers were asked to do the finger movements while keeping the wrist angle at one of 6 specified positions, including four angles on the up-down axis of the wrist, bending the wrist to the right, and keeping it straight. 


Due to different data acquisition rates of the Myo armband and Leap Motion there is a need for data synchronization and alignment. Global alignment was achieved by matching the dataframes from the two devices, minimizing the difference between the time stamps of the individual dataframes, with a maximum difference of 10 ms.

After alignment, the sEMG  and limb angle data was filtered with a low pass filter with a frequency of 10 Hz and 4Hz respectively to achieve noise reduction. The sEMG data was split into windows of size 128, which formed the input vector for the recurrent neural network.

\section{Experiment Setup}

In this section we provide the design parameters of the two RNN approaches used in our experiments. The first approach represents standard RNN technology, while the second approach incorporates optimizations selected by us for the task of continuous gesture recognition. 

Adversarial domain adaptation was implemented with a gradient reversal layer after the feature generation block. The discriminator block consists of 2 fully connected layers with 256 neurons in the first layer and 15 or 18 neurons in the second (output) layer, since there are 15 angles to predict in the case of an immobile wrist and 18 in case of a mobile wrist.

The hyper-parameters of the recurrent neural network with GRU used in our experiments are: 

\begin{itemize}
    \item 2 GRU recurrent layers with 256 neurons each. Only the last output was fed to the next layer.
    \item 2 fully connected layers with 256 and 15 (18) neurons respectively.
    \item Optimizer: Adam \cite{kingma2014adam} with a learning rate of 0.001.
    \item Trained for 30 epochs or until there was no validation gain during 8 epochs.
\end{itemize}

The hyper-parameters for the recurrent neural network with SRU are: 

\begin{itemize}
    \item 2 SRU recurrent layers with 256 neurons each.
    \item Global Average Pooling layer.
    \item 2 fully connected layers with 256 and 15 (18) neurons respectively.
    \item Optimizer: Adam with learning rate equals to 0.001.
    \item Net was trained for 30 epochs or until there was no validation gain during 8 epochs.
\end{itemize}
    

\section{Evaluation Metrics}

\makeatletter
\DeclareRobustCommand{\vardivision}{%
  \mathbin{\mathpalette\@vardivision\relax}%
}
\newcommand{\@vardivision}[2]{%
  \reflectbox{$\m@th\smallsetminus$}%
}
\makeatother

In our evaluation, we measure the root mean square error (RMSE) and normalized root mean square error (NRMSE).  RMSE is used as a standard way to measure regression error. In addition, we used NRMSE to normalize the errors to account for different limb movement ranges. RMSE and NRMSE are given in equations~\ref{rmse} and~\ref{nrmse} respectively. 

$y_i$ denote the real values and $\hat{y_i}$ the predicted values. $y_{range}$ is the range of real values. $\Theta_i = y_i \vardivision y_{range}$ and $\hat{\Theta_i} = \hat{y_i} \vardivision y_{range}$. 

where $n$ is the number of angles predicted (15 for an immobile wrist and 18 for a mobile wrist).

\begin{equation}
\label{rmse}
    RMSE = \sqrt[]{\frac{1}{n}\sum^{n}_{i=0}{(y_i - \hat{y_i})^2}}
\end{equation}

\begin{equation}
\label{nrmse}
    NRMSE = \sqrt[]{\frac{1}{n}\sum^{n}_{i=0}{(\Theta_i - \hat{\Theta_i})^2}}
\end{equation}

\newcolumntype{C}[1]{>{\centering\let\newline\\\arraybackslash\hspace{0pt}}m{#1}}


\begin{table*}[tb]
\centering
\caption{Results for immobile wrist}
\label{all_table_static}
\begin{tabular}{|l|p{32mm}|c||C{20mm}|C{20mm}||C{20mm}|C{20mm}|}
\hline
\multirow{2}{10mm}{Metric} & \multirow{2}{12mm}{Model} & \multirow{2}{15mm}{Intra session} & \multicolumn{2}{p{16mm}||}{Inter session} & \multicolumn{2}{p{16mm}|}{Inter subjects} \\ \cline{4-7} 
& & & no ADA & with ADA & no ADA & with ADA \\ 
\hline
\multirow{3}{*}{RMSE} 
                  & Gaussian Process & 23.29$\pm$0.15 & 23.44$\pm$0.05	& -	 & 23.91$\pm$0.01 & - \\ \cline{2-7} 
                  & GRU Recurrent neural net & 18.26$\pm$0.04	& 19.80$\pm$0.14 &	20.50$\pm$0.07	& 23.45$\pm$0.30 & \textbf{22.29$\pm$0.24} \\ \cline{2-7} 
                  & SRU Recurrent neural net & \textbf{17.99$\pm$0.04} & \textbf{19.24$\pm$0.04} &	19.70$\pm$0.03	& 23.14$\pm$0.06 &	\textbf{22.07$\pm$0.18} \\ 
                  \cline{2-7} 
                  \hline \hline
\multirow{3}{*}{NRMSE} 
                  & Gaussian Process & 0.2456$\pm$0.0016  & 0.1924$\pm$0.0000 & - &	0.2034$\pm$0.0001 &	- \\ \cline{2-7} 
                  & GRU Recurrent neural net & 0.1552$\pm$0.0004 & 0.1604$\pm$0.0008 & 0.1684$\pm$0.0005 & 0.1984$\pm$0.0023 & \textbf{0.1890$\pm$0.0021} \\ \cline{2-7} 
                  & SRU Recurrent neural net & \textbf{0.1512$\pm$0.0004} &	\textbf{0.1558$\pm$0.0004} &	0.1613$\pm$0.0004 &	0.1967$\pm$0.0002 &	\textbf{0.1879$\pm$0.0019} \\ 
                  \cline{2-7} 
                  \hline
\end{tabular}
\end{table*}

\begin{table*}[tb]
\centering
\caption{Results for mobile wrist}
\label{all_table_wrist}
\begin{tabular}{|l|p{32mm}|c||C{20mm}|C{20mm}||C{20mm}|C{20mm}|}
\hline
\multirow{2}{10mm}{Metric} & \multirow{2}{12mm}{Model} & \multirow{2}{15mm}{Intra session} & \multicolumn{2}{p{16mm}||}{Inter session} & \multicolumn{2}{p{16mm}|}{Inter subjects} \\ \cline{4-7} 
& & & no ADA & with ADA & no ADA & with ADA \\  
\hline
\multirow{3}{*}{RMSE}
                  & Gaussian Process & 22.33$\pm$0.24 &	22.23$\pm$0.00	& - & 22.39$\pm$0.01 &	- \\ \cline{2-7} 
                  & GRU Recurrent neural net & 18.60$\pm$0.05 & 19.19$\pm$0.05 & 19.93$\pm$0.06 & 21.74$\pm$0.20 & \textbf{20.94$\pm$0.17} \\ \cline{2-7} 
                  &  SRU Recurrent neural net  & \textbf{18.26$\pm$0.11} & \textbf{18.83$\pm$0.04} & 19.55$\pm$0.03 & 21.50$\pm$0.11 & \textbf{20.89$\pm$0.09} \\ \cline{2-7} 
                  \hline \hline
        
\multirow{3}{*}{NRMSE} 
                  & Gaussian Process &  0.2360$\pm$0.0033	& 0.1770$\pm$0.0000 & - & 0.1824$\pm$0.0001 & - \\ \cline{2-7} 
                  & GRU Recurrent neural net &  0.1558$\pm$0.0004 & 0.1518$\pm$0.0004 & 0.1580$\pm$0.0006 & 0.1757$\pm$0.0016 & \textbf{0.1698$\pm$0.0012} \\ \cline{2-7} 
                  & SRU Recurrent neural net & \textbf{0.1516$\pm$0.0013} & \textbf{0.1490$\pm$0.0000} & 0.1550$\pm$0.0000 & 0.1742$\pm$0.0008 & \textbf{0.1695$\pm$0.0007} \\
                  
                  \cline{2-7} 
                  \hline
\end{tabular}
\end{table*}

\section{Evaluation} \label{sec:evaluation}

In order to show the superiority of the improved RNN approach (RNN with SRU) over other recent approaches from the literature, namely a standard RNN with GRU approach (similar to \cite{Du2017}) and Gaussian Process (as in \cite{Quivira2018}), we applied these three methods to our datasets with and without mobile wrists.
In addition, we applied adversarial domain adaption (ADA) to the RNN approaches to evaluate the transfer ability of each of these methods. Note, that ADA is not applicable to Gaussian Processes.   

We split the data into training, validation and test sets as follows: 

\begin{enumerate}
    \item Each session was split into blocks of 12 seconds each.
    \item A period of 3 seconds was randomly sampled from each block.
    \item Half of these periods formed a validation set, while the other half formed the test set 
    \item The training set was created from the sessions after removing all periods that were used for the validation and test set. 
\end{enumerate}

In other words, to get intra-session results all training, validation and test sets from all session were combined together. Half of this data was used for training, 25\% was used for validation and the remaining 25\% for testing. 

In order to obtain inter-session results, we split the data set into five partitions, where the partitions do not share data from the same session. We used four of the partitions for training and validation, and the remaining partition for testing. This was repeated five times where each partition is used as a test set exactly once. 

A similar approach was used to evaluate inter-subject performance, where each of the five partitions contained data from a single person.

\section{Results}

The results of the evaluation with immobile wrist and mobile wrist data are shown in tables~\ref{all_table_static} and~\ref{all_table_wrist} respectively. The evaluation results for the three machine learning algorithms are presented in terms of RMSE and NRMSE for the three types of experiments: intra-session, inter-session, and inter-subject (as described in Section~\ref{sec:evaluation}). For inter-session and inter-subject we also compare results with and without the application of adversarial domain adaptation (ADA).

First of all, recurrent neural networks with SRUs show better performance in most cases, and comparable performance in the case of inter-subject experiments. These results hold for both mobile and immobile wrists. The apparent gesture recognition accuracy improvement when comparing mobile wrist with immobile wrist results is due to an increased number of predicted angles in the mobile wrist case (18 vs. 15). Therefore, the two cases are not really comparable to each other. 

As can be seen, adversarial domain adaptation (ADA) improves inter-subject accuracy, while actually showing worse results for inter-session experiments. This demonstrates that inter-session differences are not significant enough for the inhibition of session-specific features (and the resulting prevention of over-fitting) offered by ADA to have a positive impact on accuracy. However the case is different for inter-subject testing, where the difference in recorded data can be quite substantial.

\section{Conclusion and Outlook}

In this paper, we presented experiments with three machine learning approaches applied to continuous gesture recognition based on sEMG measurements. Our main contributions are: 

\begin{enumerate}
    \item We presented for the first time an evaluation of machine learning techniques with mobile wrists, which significantly enhance the functionality of artificial limbs.
    \item We demonstrated that RNNs with SRUs are superior to other state-of-the-art approaches to gesture recognition proposed in the literature, specifically Gaussian Processes and RNNs with GRUs. 
    \item We presented for the first time inter-subject gesture recognition experiments and showed the advantages of adversarial domain adaption for this task. 
\end{enumerate}

In future work, we intend to further investigate the case of inter-subject gesture recognition, including the development of improved and customized domain adaptation techniques. Also, we plan to look into the application of Bayesian methods to further improve continuous gesture recognition accuracy. 

\bibliography{main}
\bibliographystyle{IEEEtran}

\end{document}